\documentclass{article}

\PassOptionsToPackage{numbers, compress}{natbib}
\usepackage[preprint]{neurips_2024}

\usepackage[utf8]{inputenc} 
\usepackage[T1]{fontenc}    
\usepackage{hyperref}       
\usepackage{url}            
\usepackage{booktabs}       
\usepackage{amsfonts}       
\usepackage{nicefrac}       
\usepackage{microtype}      
\usepackage{xcolor}         
\usepackage{graphicx}
\usepackage{amsmath}
\usepackage{amssymb}
\usepackage{subcaption}
\usepackage{wrapfig}
\usepackage{pifont}
\usepackage{algorithmic}
\usepackage[ruled]{algorithm2e}
\usepackage{multirow}
\usepackage{fontawesome}
\usepackage{marvosym}

\newcommand{\cmark}{\ding{51}}
\newcommand{\xmark}{\ding{55}}

\newcommand{\ours}{HIA}

\newcommand{\ie}{\emph{i.e.}}

\newcommand{\etc}{\emph{etc}}

\definecolor{pearThree}{HTML}{E74C3C}
\definecolor{pearDark}{HTML}{2980B9}
\definecolor{pearDarker}{HTML}{1D2DEC}

\hypersetup{
	colorlinks,
	citecolor=pearDark,
	linkcolor=pearThree,
    breaklinks=true,
	urlcolor=pearDarker}

\title{High Quality Human Image Animation using Regional Supervision and Motion Blur Condition}

\author{%
  Zhongcong Xu$^{1}$\thanks{Equal contributions.}
  \quad Chaoyue Song$^{2*}$\quad Guoxian Song$^{3*}$\quad Jianfeng Zhang$^3$\quad Jun Hao Liew$^3$\\\textbf{Hongyi Xu$^3$\quad You Xie$^3$\quad Linjie Luo$^3$\quad Guosheng Lin$^2$ \quad Jiashi Feng$^3$ \quad Mike Zheng Shou$^1$\thanks{Corresponding author.}}\\
  $^1$Showlab, National University of Singapore \quad$^2$Nanyang Technological University\quad$^3$ByteDance\\
  \texttt{zhongcongxu@u.nus.edu\quad Chaoyue002@e.ntu.edu.sg\quad guoxiansong@bytedance.com}
}

\begin{document}

\maketitle

\begin{abstract}
Recent advances in video diffusion models have enabled realistic and controllable human image animation with temporal coherence. Although generating reasonable results, existing methods often overlook the need for regional supervision in crucial areas such as the face and hands, and neglect the explicit modeling for motion blur, leading to unrealistic low-quality synthesis. To address these limitations, we first leverage regional supervision for detailed regions to enhance face and hand faithfulness. Second, we model the motion blur explicitly to further improve the appearance quality. Third, we explore novel training strategies for high-resolution human animation to improve the overall fidelity. Experimental results demonstrate that our proposed method outperforms state-of-the-art approaches, achieving significant improvements upon the strongest baseline by more than 21.0\% and 57.4\% in terms of reconstruction precision (L1) and perceptual quality (FVD) on HumanDance dataset. Code and model will be made available.
\end{abstract}

\section{Introduction}

Human image animation, the process of animating a static reference image according to a prescribed motion signal, holds immense potential for creating highly realistic and adaptable experiences in fields such as entertainment, movie industry, and virtual reality.
Graphic approaches~\cite{collet2015high,xiang2021modeling,beeler2011high,ghosh2011multiview,guo2019relightables} create virtual avatars using template registration or multi-camera light stages and then animate the created avatars based on the provided motion signal. 
Recent efforts~\cite{siarohin2019first,wang2021one,geng20193d,chan2019everybody,xu2023xagen,zhang2023avatargen,hong2023evad,song20213d,song2023unsupervised} investigate data-driven approaches for human avatar animation based on generative models.

Existing works for data-driven animation can be classified into two categories, \ie, GAN-based~\cite{zhao2022thin,siarohin2021motion} and diffusion-based methods~\cite{wang2023disco,karras2023dreampose}. The GAN-based works typically explore image warping based on the optical flow, while the diffusion-based works leverage the visual priors of a pre-trained diffusion model to enhance the animation quality. 
These works demonstrate the capabilities of generating unprecedented realistic animation results with long-range temporal coherence, which has spawned a wide range of downstream applications in the industry.

Despite producing plausible animation results, such methods have several drawbacks: (1) The learning objective of these works is the MSE loss for the entire body image. Though effective for training, such a straightforward learning objective cannot guarantee a promising appearance for two important yet challenging regions, \ie, face and hands. The main reason is that these parts have relatively smaller scales than arms, legs, and torso in the human body. Consequently, the supervision provided by full-body MSE loss may not effectively propagate gradients to these smaller regions, leading to suboptimal appearance quality in the face and hands. (2) Human-centric videos contain a wide range of daily activities, such as object manipulation, gesturing, dancing, \etc. Due to the rapid motion and limitations of capturing devices, motion blur is commonly present in human-centric videos, particularly in regions such as the hands. However, none of the existing works consider motion blur issues, leading to the unconditional synthesis of motion blur in animation results. (3) The default noise scheduler has been proved flawed and not suitable for high-resolution training~\cite{chen2023importance,lin2024common}. This issue also hinders the increase of training resolution for diffusion-based human image animation. 

\begin{figure}[t]
\centering
\includegraphics[width=0.95\textwidth]{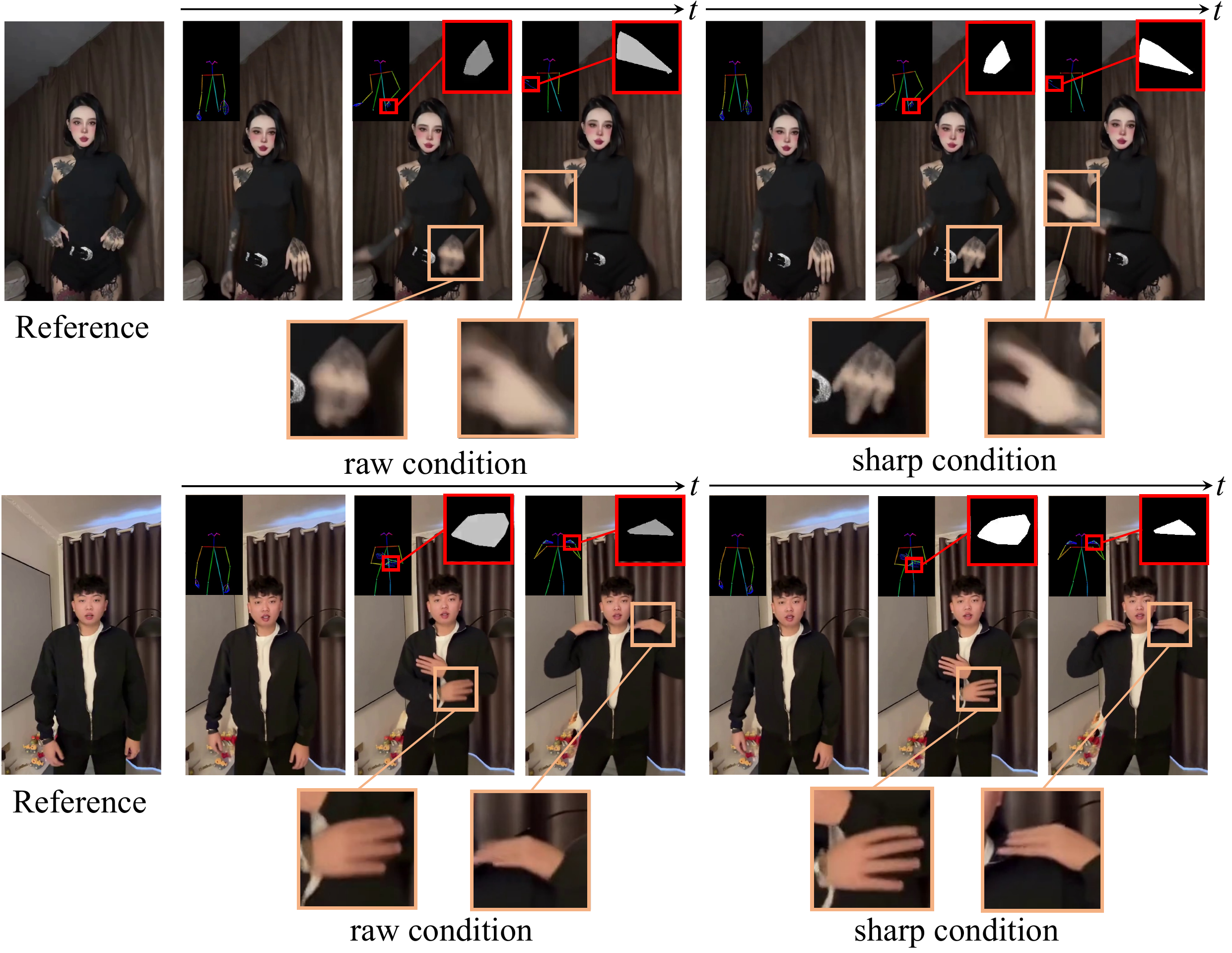}
\caption{We introduce \ours{}, a high-quality human image animation framework designed to generate realistic results, particularly for small-scale regions such as faces and hands. Our approach incorporates explicit conditioning on the motion blur of hands, enabling precise control over hand sharpness. We overlay the motion signal and motion blur condition on the top left and top right corners of each synthesized video frame respectively.}
\label{fig:teaser}
\vspace{-3mm}
\end{figure}

In this work, we aim to enhance both the overall sing-frame quality and the details of face and hands for human image animation, as shown in Figure~\ref{fig:teaser}. Thus, we propose a {\bf H}igh quality human {\bf I}mage {\bf A}nimation framework ({\bf \ours{}}). \ours{} is built upon recent diffusion-based human image animation methods. We adopt a similar architecture design. To address the aforementioned limitations of the existing works, we first propose regional supervision to ensure the faithfulness of the face and hands during training via a masked MSE loss term. We also utilize the cosine similarity loss to preserve the identity of the synthesized faces. Second, we incorporate motion blur conditioning for hands by integrating hand movement vectors and sharpness scores for each video frame with the driving signal. Third, we investigate the effects of signal-to-noise ratio (SNR) in the noise scheduler and implement a progressive training strategy for temporal modules to maintain high-quality video frames.
We conduct extensive experiments on two benchmarks, \ie, TikTok~\cite{jafarian2021learning} and a HumanDance dataset collected by ourselves, demonstrating the superiority of \ours{} over state-of-the-art methods in terms of sing-frame quality, video fidelity, and generalization ability. 
Our contributions consists of three key facets: (1) We propose a human image animation framework, marrying regional supervision, shifted SNR, and progressive training strategy, to enable high-quality image animation. (2) We are the first diffusion based work to handle the motion blur issue in human-centric videos. (3) Comprehensive experiments show that \ours{} outperforms state-of-the-art methods in both single-frame and video quality.

\section{Related work}
{\bf Diffusion models for human image animation.} The task of human image animation aims to synthesize the video of a reference identity and background following a particular motion sequence~\cite{zhao2022thin,siarohin2019first,chan2019everybody,yoon2021pose,yu2023bidirectionally}. Conventional methods for this task either choose to reconstruct the 3D human avatar first~\cite{svitov2023dinar} or learn to warp the reference image into the target pose~\cite{siarohin2021motion}. Recent advancements in the diffusion models~\cite{rombach2022high,zhang2023adding,cao2023masactrl,ye2023ip} have inspired a line of research works exploring their application in animation tasks. DreamPose~\cite{karras2023dreampose} adopts CLIP~\cite{radford2021learning} encoder to preserve the reference image and combines pose information with noisy latent noise for pose transfer.  DisCo~\cite{wang2023disco} improves upon DreamPose by using separated reference conditions for foreground and background respectively. However, these methods cannot guarantee temporal coherence because they process animation frame by frame. To alleviate this issue, the following work MagicAnimate~\cite{xu2023magicanimate} and AnimateAnyone~\cite{hu2023animate} utilizes temporal attention~\cite{guo2023animatediff} to improve the temporal consistency. Additionally, they propose UNet-based appearance encoders to better preserve the reference image. 
The most recent work Champ~\cite{zhu2024champ} shares a similar architecture design with them while utilizing SMPL~\cite{smpl} to provide a dense and robust motion sequence.

{\bf Motion guidance for human image animation.} Accurate and robust motion sequences are crucial for human image animation as they directly impact the controllability and quality of the generated content. Among all the human pose formats, 2D keypoint estimation is the most advanced, such as DWPose~\cite{yang2023effective} and RTMPose~\cite{jiang2023rtmpose}. These methods provide more expressive keypoints than openpose~\cite{cao2017realtime} and are widely used in human image animation works~\cite{hu2023animate,feng2023dreamoving}. Though providing stable control signal, 2D keypoints are too sparse because they only focus on the major joints in the human body, face, and hand. Therefore, several works~\cite{xu2023magicanimate,chen2024wear} adopt DensePose~\cite{guler2018densepose} as pose guidance to animate human images or change garments for virtual try-on. In addition to these pixelwise motion sequences, statistical parametric models, such as SMPL~\cite{smpl}, can provide 3D vertices for naked human body surface, which can also serve as pose guidance~\cite{zhu2024champ,zhang2023avatargen}. However, SMPL has limitations in modeling detailed regions such as facial expressions and hand poses. Thus, another line of works leverages expressive parametric model, \ie, SMPL-X~\cite{SMPL-X}, to implement human image animation~\cite{xu2023xagen,svitov2023dinar}. In this work, we choose 2D keypoints as our driving signal, while we also observed that this sparse joint condition cannot encode the motion speed and motion blur of the human-centric videos. To address this, we propose incorporating human movement and hand sharpness scores to model motion blur more effectively.


\section{Method}
In this section, we introduce \ours{}, a human image animation framework equipped with regional supervision, explicit motion blur condition, as well as carefully designed training strategies. \ours{} enables high-quality human image animation with realistic faces and hands.

\begin{figure}[t]
\vspace{-1em}
\centering
\includegraphics[width=\textwidth]{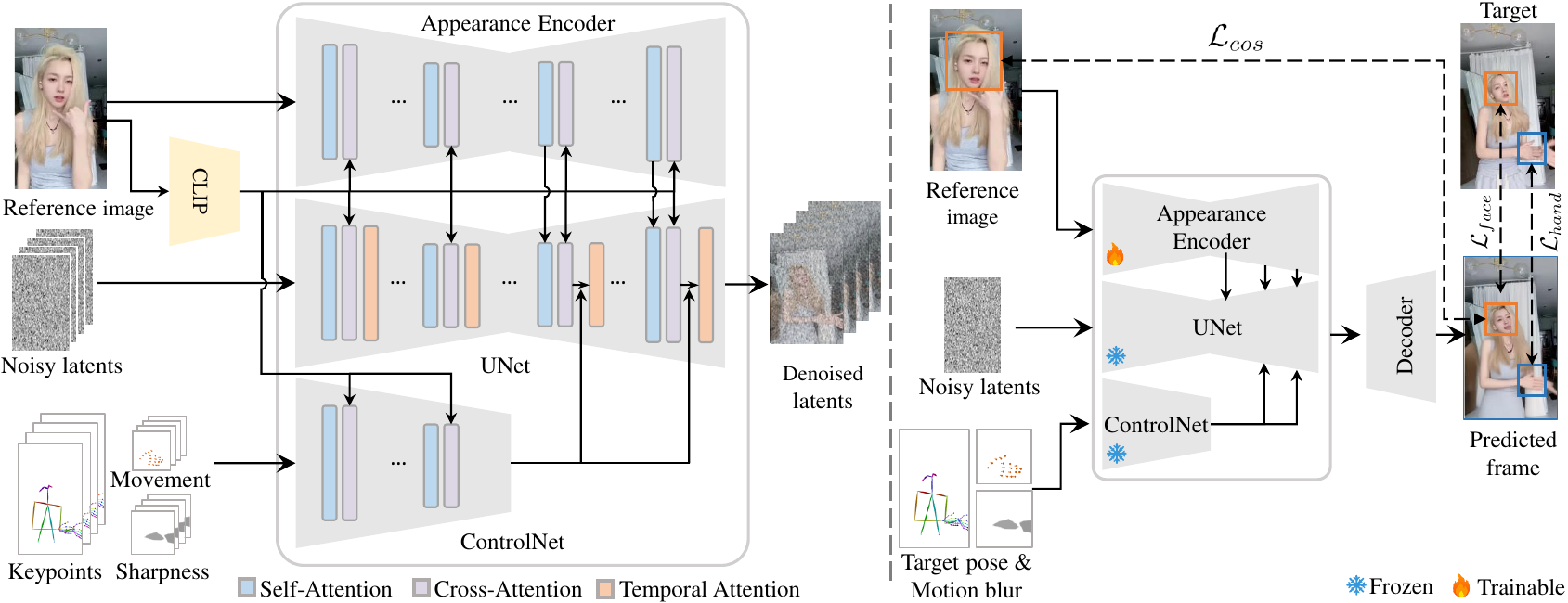}
\vspace{-3mm}
\caption{
Given a random noisy latent, a reference image, a motion sequence, and motion blur condition, our model synthesizes the avatar using the identity and background from the reference image and animates the avatar adhering to the provided motion sequence (\textbf{left panel}). To enhance the quality of the face and hands, we devise a regional supervision stage that fine-tunes appearance encoder with MSE and cosine similarity loss terms ({\bf right panel}). 
}
\label{fig:network}
\end{figure}

Given a reference image \(\mathbf{I}_\text{ref}\) and a driving signal \(\boldsymbol{p}^{1:N}\), where \(N\) is the motion length, the goal of HIA is to synthesize a human-centric video that maintains the character appearance and background of \(\mathbf{I}_\text{ref}\) while adhering to the motion represented by \(\boldsymbol{p}^{1:N}\). To achieve this, we follow the prior works~\cite{xu2023magicanimate, hu2023animate, zhu2024champ} and design a framework consists of UNet-based appearance encoder, CLIP encoder, UNet, and ControlNet, as depicted in Figure~\ref{fig:network}. We train the model in two stages, with the first stage for spatial modules and the second stage for temporal modules.
Our proposed method not only aims to generate realistic motion but also enhances details in small-scale regions like the face and hands. To achieve this, in addition to the two standard training stages, we introduce an additional regional supervision stage (Sec.~\ref{sec:region}), as shown in the right panel of Figure~\ref{fig:network}. This stage focuses on improving the quality of details in regions such as the face and hands, thereby enhancing the overall realism of the generated videos.

Moreover, due to the rapid articulated motion of human body and the limitations of capturing devices, motion blur is ubiquitous in human-centric videos, such as TikTok~\cite{jafarian2021learning} dancing videos. However, all of the prior works neglect this factor and none of them model the motion blur explicitly. As a result, these approaches, when trained on human-centric datasets, inherently learn to generate blurry results unconditionally. This issue is particularly pronounced in the hand region, as hands are the end parts of the human body skeleton and human-centric videos contain a significant amount of gestures and hand movements. HIA addresses this challenge by explicitly modeling hand motion blur (Sec.~\ref{sec:blur}). Specifically, we compute the hand movement vector \(\boldsymbol{v}\) using the hand keypoints from two consecutive frames. In addition, we crop the hand images and compute the sharpness score \(\boldsymbol{s}\) as conditional signals. These hand movement vectors and sharpness scores are then fed into our framework, enhancing the clarity of hands in the generated videos.

Existing human image animation methods either train the video diffusion model using epsilon prediction at low resolution~\cite{xu2023magicanimate} or adopt velocity prediction to stabilize the training process~\cite{zhu2024champ}, which decreases video sharpness. We argue that these approaches are not suitable for training high-resolution animation models due to the limitations of their noise schedulers. To alleviate this issue, we employ a shifted signal-to-noise ratio (SNR) technique. Additionally, we design a progressive training strategy (Sec.~\ref{sec:training}) to further improve temporal coherence and maintain spatial quality.

\subsection{Regional supervision}
\label{sec:region}
Enhancing details in small-scale regions such as the face and hands is a challenging yet important issue in avatar generation \cite{xu2023xagen, zhang2023avatargen} and reconstruction \cite{song2023moda, yang2022banmo}. To tackle this challenge, we introduce regional supervision. Specifically, in addition to the main training stages on spatial and temporal modules, we implement an additional fine-tuning stage that incorporates regional supervision to improve these detailed areas.

Given the target image $\mathbf{I}_{tgt}$ and the predicted frame $\mathbf{I}_{pre}$, our objective is to ensure that the face and hands in $\mathbf{I}_{pre}$ closely resemble those in $\mathbf{I}_{tgt}$. To achieve this, we first crop the face and hands using masks $\mathbf{M}_{face}$ and $\mathbf{M}_{hand}$. We then calculate the regional MSE losses as follows,
\begin{equation}
    \mathcal{L}_{face} = \frac{\sum \left\|(\mathbf{I}_{tgt} - \mathbf{I}_{pre})\odot\mathbf{M}_{face}\right\|^{2}_{2}} {\sum \mathbf{M}_{face}}, \quad \mathcal{L}_{hand} = \frac{\sum \left\|(\mathbf{I}_{tgt} - \mathbf{I}_{pre})\odot\mathbf{M}_{hand}\right\|^{2}_{2}} {\sum \mathbf{M}_{hand}}, 
    \label{loss_mse}
\end{equation}
where $\odot$ denotes Hadamard product, and we calculate $\mathcal{L}_{hand}$ for both hands. Additionally, we encourage the similarity between the face in the reference image $\mathbf{I}_{ref}$ and the predicted frame $\mathbf{I}_{pre}$ by calculating a face cosine similarity loss. We use Insightface \cite{deng2019arcface} to extract the face embeddings $\boldsymbol{\psi}_{ref} \in \mathbb{R}^{512}$ and $\boldsymbol{\psi}_{pre} \in \mathbb{R}^{512}$, which are then used to calculate the cosine similarity loss,
\begin{equation}
\mathcal{L}_{cos} = 1 -
\frac{\boldsymbol{\psi}_{ref} \cdot \boldsymbol{\psi}_{pre}}{\left\|\boldsymbol{\psi}_{ref}\right\|\left\|\boldsymbol{\psi}_{pre}\right\|}.
    \label{cos_loss}
\end{equation}
We incorporate these regional losses only in the regional supervision stage which is to fine-tune the spatial modules after the spatial stage, please refer to more details in Sec.~\ref{sec:training}.

\subsection{Motion blur condition}
\label{sec:blur}
Human-centric videos contain diverse human activities such as talking and dancing. It is common to observe abundant motion blur in these daily activities. Without explicit modeling, prior works learn to generate these ubiquitous motion blur, leading to unrealistic video results. To improve the generation quality, we propose a motion blur conditioning approach.

The motion blur in a dancing video is reflected as a blurry region. It is usually caused by the rapid motion of hands. To compute the conditioning signal for motion blur, we process each video in the dataset from two perspectives, \ie, hand sharpness scores and hand movement vectors. To measure the hand sharpness, we crop the hand images \(\mathbf{I}_\text{h}\) from each video frame and then apply a Laplacian filter to compute the second derivative
\begin{equation}
    Laplace(\mathbf{I}_\text{h}) = \frac{\partial^2 \mathbf{I}_\text{h}}{\partial x^2} + \frac{\partial^2 \mathbf{I}_\text{h}}{\partial y^2},
\end{equation}
where \(x\) and \(y\) are columns and rows of image pixel. We further calculate the variance of the Laplacian operator to get the sharpness score \(\boldsymbol{s}\).
In addition, HIA uses 2D keypoint sequence as the driving signal. For each training video, we estimate the keypoint sequence frame by frame and get \(\boldsymbol{p}^{1:N}\). Based on \(\boldsymbol{p}^{1:N}\), we compute the movement vector \(\boldsymbol{v}=\boldsymbol{p}_\text{h}^{i}-\boldsymbol{p}_\text{h}^{i-1}\) for the hands in each frame at timestep \(i\), where \(\boldsymbol{p}_\text{h}\) denotes the hand keypoints. 

To condition HIA on the above motion blur conditions, we overlay the motion vector \(\boldsymbol{v}\) and sharpness score \(\boldsymbol{s}\) on the hand regions of the openpose keypoint sequence. In particular, we compute the average values for the driving signals and input it into the ControlNet in HIA.

\subsection{Training}
\label{sec:training}
\begin{wrapfigure}{r}{6.9cm}
\vspace{-5mm}
\begin{algorithm}[H]
\SetAlgoLined

\textbf{Require} $0<\gamma<1$\;
$\beta = \{\beta_t\}, \alpha = \{\alpha_t\}, t \in \{0...T\}$\;

for $t~\text{in}~\{0...T\}$ do

\quad$\beta_t \leftarrow 0.00085*(1-\frac{t-1}{T-1}) + 0.012*\frac{t-1}{T-1}$\;

\quad$\alpha_t \leftarrow 1-\beta_t$\;

end for

$snr = \{snr_t\}$

for $t~\text{in}~\{0...T\}$ do

\quad$snr_t\leftarrow \gamma*\prod_{i=0}^{t}\alpha_i / (1 - \prod_{i=0}^{t}\alpha_i)$\;

end for

for $t~\text{in}~\{1...T\}$ do

\quad$\alpha_t^\text{c} \leftarrow snr_t/(1+snr_t)$\;

\quad$\alpha_{t-1}^\text{c} \leftarrow snr_{t-1}/(1+snr_{t-1})$\;

\quad$\beta_t \leftarrow 1- \alpha_t^\text{c}/\alpha_{t-1}^\text{c}$\;
end for

\textbf{Return} $\beta$
 \caption{Shift of the signal-to-noise ratio.}
 \label{alg:shift_snr}
\end{algorithm}
\vspace{-5mm}
\end{wrapfigure}
Following the training convention of plug-and-play video diffusion modules like Animatediff~\cite{guo2023animatediff}, existing works train spatial and temporal modules sequentially in independent stages. HIA also follows this convention and trains spatial module, \ie, appearance encoder, ControlNet, and base UNet, in the first stage. Then we fine-tune the spatial modules with the regional supervision stage, where we optimize the identity preservation ability for details like face and hands. Finally, we insert the temporal attention layers and train these temporal layers only.

\textbf{Shift SNR.} Different from prior works~\cite{xu2023magicanimate,salimans2022progressive} which utilize the default noise scheduler, we empirically find that the default scheduler cannot work well for higher resolution, such as 512\(\times\)896. The reason is that this noise scheduler cannot destroy the ground truth image in the forward process when the training resolution is high~\cite{chen2023importance,lin2024common}. Thus, we adjust the SNR of the scheduler during training. Specifically, as shown in Algorithm~\ref{alg:shift_snr}, we first compute the SNR based on linear scheduler and then reduce the SNR by a factor \(\gamma\), where \(0<\gamma<1\). We then employ the \(\beta\) derived from the shifted SNR for training.

\textbf{Regional supervision stage.} After training the spatial modules in the first stage, we fine-tune them with the regional supervision stage to improve the identity preservation ability of face and hands. To obtain clear denoised images for calculating the regional losses, we add noise with a small timestep during this stage. According to the observation in ReFL \cite{xu2024imagereward}, we randomly sample the noise with a timestep range from 0 to 124, rather than 0 to 999 when training UNet in the first stage. We then directly predict the denoised latent $x_{0}^{\prime}$ with one step, which is clear enough to compare with the target image. In this stage, we freeze the UNet and ControlNet, and only fine-tune the appearance encoder to avoid the impact of timestep restrictions on the UNet.

\textbf{Progressive training.} Existing works choose to freeze spatial modules in the temporal training stage since the spatial layers are already capable to generate nearly coherent frames. Ideally, the trained temporal attention layers serve to smooth the frame sequences without impacting the spatial content. However, in practice, we notice that the temporal layers learn appearance-relevant information, causing degradation in spatial quality. To alleviate this, we devise a progressive training strategy. We divide the temporal module training into two sub-stages. In the first stage, we train the temporal module using half resolution. While in the second stage, we train the temporal module on full resolution but we sample static images for augmentation following MagicAnimate~\cite{xu2023magicanimate}, which helps maintain the high-quality video frames generated by the spatial module.

\begin{table*}[t]
 \small
 \caption{Quantitative comparisons with baselines, with the best results highlighted in {\bf bold}.
  }
  \centering
    \begin{subtable}[t]{\linewidth}
    \caption{Quantitative comparisons on HumanDance dataset.}
    \centering
  \begin{tabular}{lccccccc}
    \toprule
    \multirow{2}{*}{Method~~}&\multicolumn{5}{c}{Image} &\multicolumn{2}{c}{Video}\\
    \cmidrule(r){2-6} \cmidrule(r){7-8} & L1\(\downarrow\)~~ & PSNR\(\uparrow\)~~  &SSIM\(\uparrow\)~~ &LPIPS\(\downarrow\)~~ & FID\(\downarrow\)~~ & FID-VID\(\downarrow\)~~& FVD\(\downarrow\)\\
    \midrule
    MagicAnimate~\cite{xu2023magicanimate}~~ &{8.95E-05}~~ &13.80~~ &{0.664}~~ &{0.132}~~ &{39.19}~~ &{61.69}~~ &{331.40} \\
    AnimateAnyone~\cite{hu2023animate}~~ &4.01E-05~~ &18.23~~ &{0.741}~~ &0.102~~ &26.41~~ &{14.85}~~ &{114.90} \\
    Champ~\cite{zhu2024champ}~~ &{3.77E-05}~~ &{19.00}~~ &0.740~~ &{0.094}~~ &{21.34}~~ &16.03~~ &118.18 \\    
    \ours{} (Ours)~~ &{\bf 2.98E-05}~~ &{\bf 20.45}~~ &{\bf 0.799}~~ &{\bf 0.074}~~ &{\bf 17.84}~~ &{\bf 4.91}~~ &{\bf 50.33} \\
    \bottomrule
  \end{tabular}
  \label{tab:comp:humandance}
  \end{subtable}
  
  \begin{subtable}[t]{\linewidth}
  \vspace{1em}
  \caption{Quantitative comparisons on TikTok~\cite{jafarian2021learning} dataset.}
  \centering
  \begin{tabular}{lccccccc}
    \toprule
    \multirow{2}{*}{Method~~}&\multicolumn{5}{c}{Image} &\multicolumn{2}{c}{Video}\\
    \cmidrule(r){2-6} \cmidrule(r){7-8} & L1\(\downarrow\)~~ & PSNR\(\uparrow\)~~  &SSIM\(\uparrow\)~~ &LPIPS\(\downarrow\)~~ & FID\(\downarrow\)~~ & FID-VID\(\downarrow\)~~& FVD\(\downarrow\)\\
    \midrule
    MagicAnimate~\cite{xu2023magicanimate}~~ &{6.62E-05}~~ &15.50~~ &{0.691}~~ &{0.157}~~ &{36.27}~~ &{36.13}~~ &{226.98} \\
    AnimateAnyone~\cite{hu2023animate}~~ &5.69E-05~~ &15.54~~ &0.674~~ &0.165~~ &41.54~~ &{25.07}~~ &{221.66} \\
    Champ~\cite{zhu2024champ}~~ &{5.30E-05}~~ &{16.49}~~ &0.690~~ &{0.143}~~ &{32.53}~~ &25.86~~ &237.96 \\
    \ours{} (Ours)~~ &{\bf 5.10E-05}~~ &{\bf 16.88}~~ &{\bf 0.720}~~ &{\bf 0.135}~~ &{\bf 30.08}~~ &{\bf 17.85}~~ &{\bf 125.91} \\
    \bottomrule
  \end{tabular}
  \label{tab:comp:tiktok}
  \end{subtable}
  
  \label{tab:comp}
\end{table*}

\subsection{Inference}
\label{sec:inference}

During inference, to improve generation stability and avoid background jittering, we diverge from previous methods that sample initial noise features from pure Gaussian noise. Instead, we encode the reference image $\mathbf{I}_{ref}$ into latent features using a VAE encoder and diffuse these latent features 999 times, which we term as \textbf{initial reference noise} to serve as the starting latent $x_T$. This small indicative bias can help UNet correctly retain the reference's background. To enhance image quality and reduce undesired artifacts, we introduce a new Classifier-Free Guidance formulation called \textbf{animation-cfg} and incorporate it into our animation denoising step $\epsilon$. Specifically, we use the reference image $\mathbf{I}_{ref}$ and motion signal $\boldsymbol{p}^{1:N}$ as control conditions, omitting them for unconditional generation. The equation is formulated as
\begin{equation}
    \begin{aligned}
        \hat{\epsilon}(x_t,t,\mathbf{I}_{ref},\boldsymbol{p}^{1:N}) = \epsilon(x_t,t,\varnothing,\varnothing) +  \omega(\epsilon(x_t,t,\varnothing,\varnothing) - \epsilon(x_t,t,\mathbf{I}_{ref},\boldsymbol{p}^{1:N})),
    \end{aligned}
\end{equation}
where the empty symbol indicates the corresponding control module is deactivated, and $\omega$ is a scalar parameter. For long sequence generation, we employ prompt traveling\cite{tseng2022edge} based on the autoregression method used in\cite{xu2023magicanimate} to mitigate jittering artifacts. Specifically, for each denoising step within the sliding windows of an animation sequence, we select a random offset number and shift the sliding windows accordingly. We then perform denoising on each window and average the overlaps to ensure smooth transitions.

\section{Experiments}
We evaluate the performance of \ours{} on two datasets: a dataset collected by ourselves named HumanDance and TikTok~\cite{jafarian2021learning}. These datasets contain diverse human dancing videos. HumanDance consists of 3,802 video clips for training and 50 videos for testing. For TikTok, we use 300 videos for training and 41 videos for testing. For each video, we process it to obtain 2D OpenPose sequences, hand movement vectors, and hand sharpness scores. Please refer to the Appendix for more details.
\subsection{Comparisons}
{\bf Baselines.}
We compare \ours{} with three state-of-the-art diffusion-based human image animation methods: MagicAnimate~\cite{xu2023magicanimate}, AnimateAnyone~\cite{hu2023animate}, and Champ~\cite{zhu2024champ}. All these methods adopt a similar framework which consists of an appearance encoder and a pose-conditioned generation backbone with temporal attention layers. Differently, MagicAnimate employs DensePose~\cite{guler2018densepose} as motion sequence and leverages ControlNet for pose transfer, while AnimateAnyone and Champ directly concatenate pose with the initial noise. In addition, AnimateAnyone chooses OpenPose as the driving signal, and Champ utilizes SMPL~\cite{smpl}.

{\bf Evaluation metrics.} To measure the animation performance, we follow the well-established evaluation metrics adopted by existing works. We evaluate single-frame quality using L1 error, SSIM~\cite{wang2004image}, LPIPS~\cite{zhang2018unreasonable}, PSNR~\cite{hore2010image}, and FID~\cite{heusel2017gans}. For video quality, we report FID-FVD~\cite{balaji2019conditional} and FVD~\cite{unterthiner2018towards}. 
Please refer to the Appendix for more details.

{\bf Quantitative comparisons.} Table~\ref{tab:comp} summarizes the quantitative results of \ours{} and baseline methods on the HumanDance and TikTok datasets. It can be observed that MagicAnimate generates animation results with lower quality in terms of both single-frame and video because it utilizes a frozen UNet, which we believe is not suitable for high-resolution human image animation due to the resolution domain gap. AnimateAnyone and Champ yield similar results, as the main difference between these baselines is the motion sequence. These methods improve upon MagicAnimate by a large margin. Nonetheless, our method achieves state-of-the-art performance on both benchmarks. Notably, \ours{} showcases significant improvements for LPIPS (21.3\%) and FID (16.4\%) on HumanDance, demonstrating superior single-frame quality. Additionally, \ours{} improves against the strongest baseline by 57.4\% and 43.2\% in terms of FVD on HumanDance and TikTok, respectively, proving the video fidelity of \ours{}.

\begin{figure}[t]
\centering
\begin{subfigure}[b]{0.49\linewidth}
 \centering
 \includegraphics[width=\textwidth]{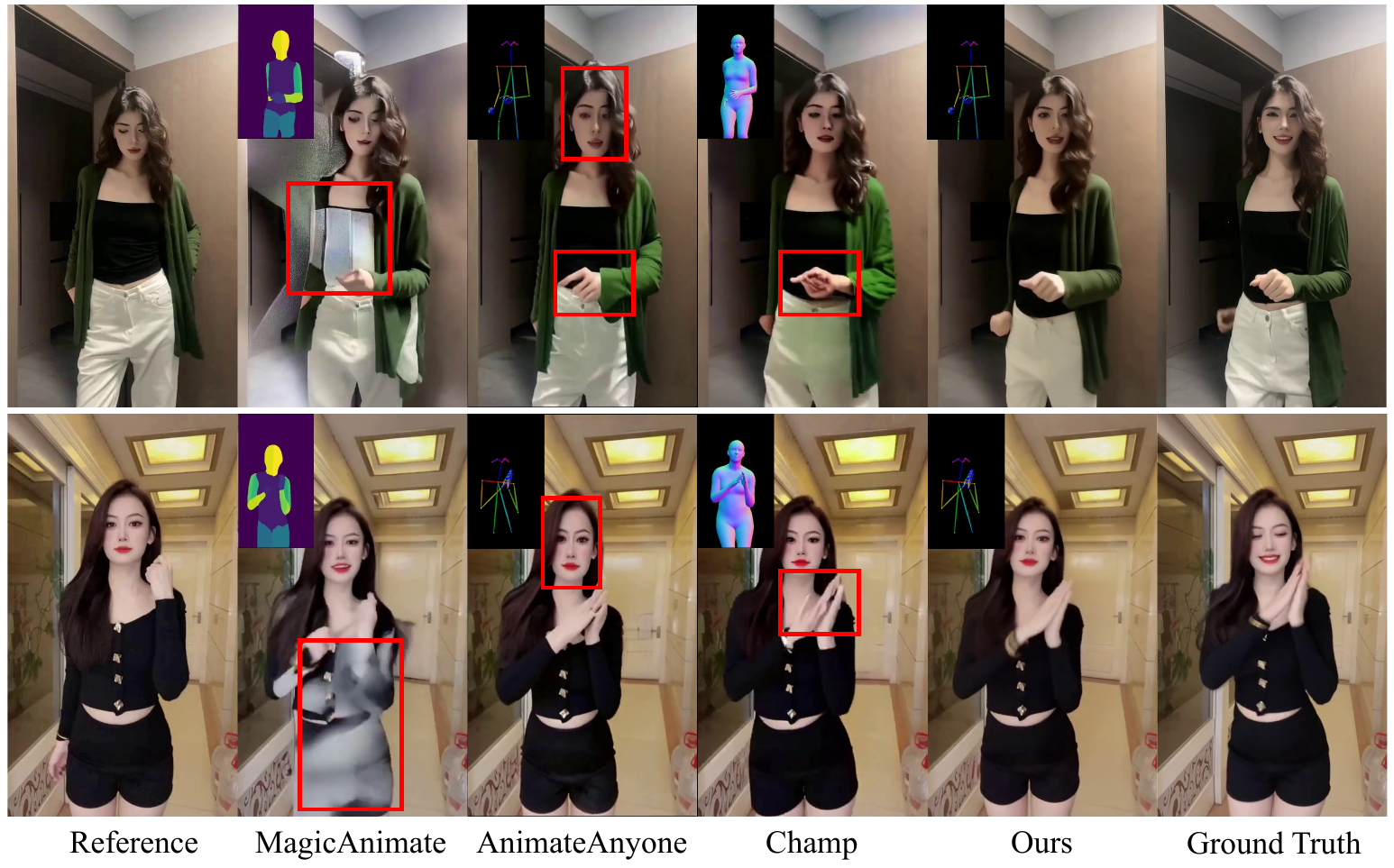}
 \caption{Comparisons on HumanDance dataset.}
 \label{fig:comp:humandance}
\end{subfigure}
\hfill
\begin{subfigure}[b]{0.49\linewidth}
 \centering
 \includegraphics[width=\textwidth]{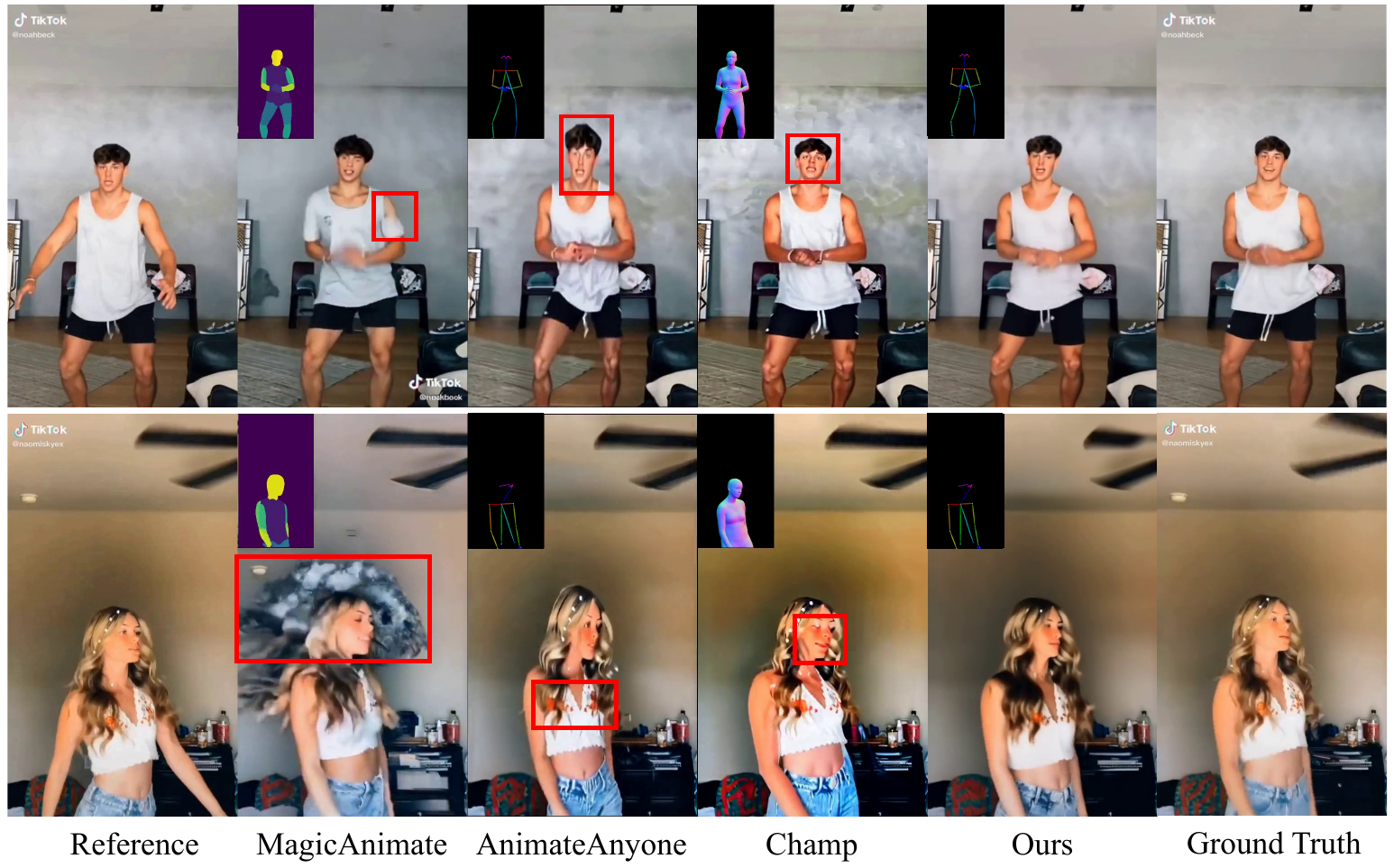}
 \caption{Comparisons on TikTok dataset.}
 \label{fig:comp:tiktok}
\end{subfigure}
\caption{Qualitative comparisons between ours and baselines on two datasets. The driving signal is overlaid in the upper left corner of each frame. Errors in the baseline methods are highlighted in red boxes. Please refer to our project page in {\it Sup. Mat.} for video results.}
\label{fig:comp}
\end{figure}

\begin{figure}[t]
\centering
\includegraphics[width=\textwidth]{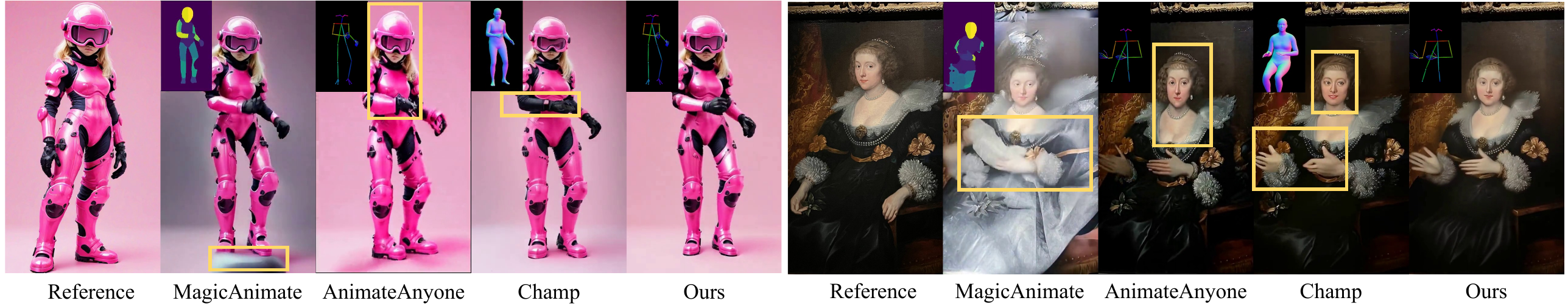}
\caption{
Qualitative comparisons between ours and baselines on unseen categories, \ie, humanoid and oil painting portraits. Errors in the baseline methods are highlighted in orange boxes. Please refer to our project page in {\it Sup. Mat.} for video results.
}
\label{fig:comp:ood}
\end{figure}

{\bf Qualitative comparisons.} In Figure~\ref{fig:comp}, we visualize the qualitative comparisons between \ours{} and the baseline methods. It can be observed that MagicAnimate synthesizes a large portion of artifacts, primarily due to the frozen UNet. AnimateAnyone and Champ generate reasonable results while the animation has high contrast color and reduces its fidelity. Additionally, their hands exhibit incorrect structure due to the lack of supervision for detailed regions. In contrast, \ours{} produces realistic animation results with clear hands and well-maintained face identity. To further evaluate generalization ability, we conducted experiments on cross-domain samples. As shown in Figure~\ref{fig:comp:ood}, \ours{} synthesizes animation results with higher quality than baselines for humanoid and oil painting portraits, demonstrating that our method has a strong generalization ability.
\subsection{Ablation studies}
\begin{table*}[t]
 \small
 \caption{Quantitative ablation studies. We evaluate the effectiveness of different components for training on the HumanDance dataset, with the best results in {\bf bold} and second best \underline{underlined}. w/o means we remove this component. \textit{Fine-tune all spatial modules} indicates that we fine-tune all spatial modules in the regional supervision stage rather than only fine-tune the appearance encoder.}
  \centering
   \resizebox{\textwidth}{!}{
  \begin{tabular}{lccccccc}
    \toprule
    \multirow{2}{*}{Method~~}&\multicolumn{5}{c}{Image} &\multicolumn{2}{c}{Video}\\
    \cmidrule(r){2-6} \cmidrule(r){7-8} & L1\(\downarrow\)~~ & PSNR\(\uparrow\)~~  &SSIM\(\uparrow\)~~ &LPIPS\(\downarrow\)~~ & FID\(\downarrow\)~~ & FID-VID\(\downarrow\)~~& FVD\(\downarrow\)\\
    \midrule
    1) w/o regional supervision ~~ &\underline{3.01E-05}~~ &\underline{20.35}~~ &{\bf 0.801}~~ &\underline{0.0751}~~ &19.26~~ &5.03~~ &57.39 \\
    2) Fine-tune all spatial modules ~~ &3.09E-05~~ &20.23~~ &0.797~~ &0.0765~~ &19.03~~ &5.42~~ &58.79 \\
    3) w/o motion blur condition ~~ &3.22E-05~~ &20.08~~ &0.797~~ &0.0771~~ &19.21~~ &\underline{5.02}~~ &57.07 \\  
    4) Default noise scheduler ~~ &3.38E-05~~ &19.42~~ & 0.787~~ &0.0796~~ &\underline{18.00}~~ &5.30~~ &\underline{56.02} \\
    
    7) Full model~~ &{\bf 2.98E-05}~~ &{\bf 20.45}~~ &\underline{0.799}~~ &{\bf 0.0740}~~ &{\bf 17.84}~~ &{\bf 4.91}~~ &{\bf 50.33} \\

    \bottomrule
  \end{tabular}
  }
   \label{tab:ablation_train}
\end{table*}

\begin{table*}[t]
 \small
 \caption{Quantitative ablation studies on inference techniques. We evaluate the effectiveness of different components on the HumanDance dataset, with the best results in {\bf bold}. A-cfg refers to animation-cfg, PT means prompt traveling, and IRN denotes initial reference noise.}
  \centering
   \resizebox{\textwidth}{!}{
  \begin{tabular}{cccccccccc}
    \toprule
    \multirow{2}{*}{A-cfg}~~& \multirow{2}{*}{PT}~~& \multirow{2}{*}{IRN}&\multicolumn{5}{c}{Image} &\multicolumn{2}{c}{Video}\\
    \cmidrule(r){4-8} \cmidrule(r){9-10} & & & L1\(\downarrow\)~~ & PSNR\(\uparrow\)~~  &SSIM\(\uparrow\)~~ &LPIPS\(\downarrow\)~~ & FID\(\downarrow\)~~ & FID-VID\(\downarrow\)~~& FVD\(\downarrow\)\\
    \midrule
    \xmark  ~~ &\xmark ~~ &\xmark ~~  &3.07E-05~~ &20.33~~ &0.797~~ &0.0745~~ &{\bf 17.54}~~ & 5.22~~ & 53.02 \\
\xmark ~~ &\cmark  ~~ &\cmark ~~ &{\bf 2.98E-05}~~ &{\bf 20.45}~~ &{\bf 0.799}~~ &{\bf 0.0740}~~ &17.85~~ & {\bf 4.90}~~ & 50.61 \\
    \cmark ~~ &\xmark  ~~ &\cmark  ~~ &{\bf 2.98E-05}~~ &20.44~~ &0.798~~ &0.0741~~ &17.69~~ & 5.01~~ & 50.78 \\
    \cmark  ~~ &\cmark  ~~ &\xmark  ~~ &3.06E-05~~ &20.34~~ &0.797~~ &0.0744~~ &17.70~~ & 5.13~~ & 52.25 \\
    \cmark  ~~ &\cmark ~~ &\cmark  ~~  &{\bf 2.98E-05}~~ &{\bf 20.45}~~ &{\bf 0.799}~~ &{\bf 0.0740}~~ &17.84~~ & 4.91~~ &{\bf 50.33} \\
    \bottomrule
  \end{tabular}
  }
   \label{tab:ablation_infer}
\end{table*}

\begin{figure}[t]
\centering
\begin{subfigure}[b]{0.48\linewidth}
 \centering
 \includegraphics[width=\textwidth]{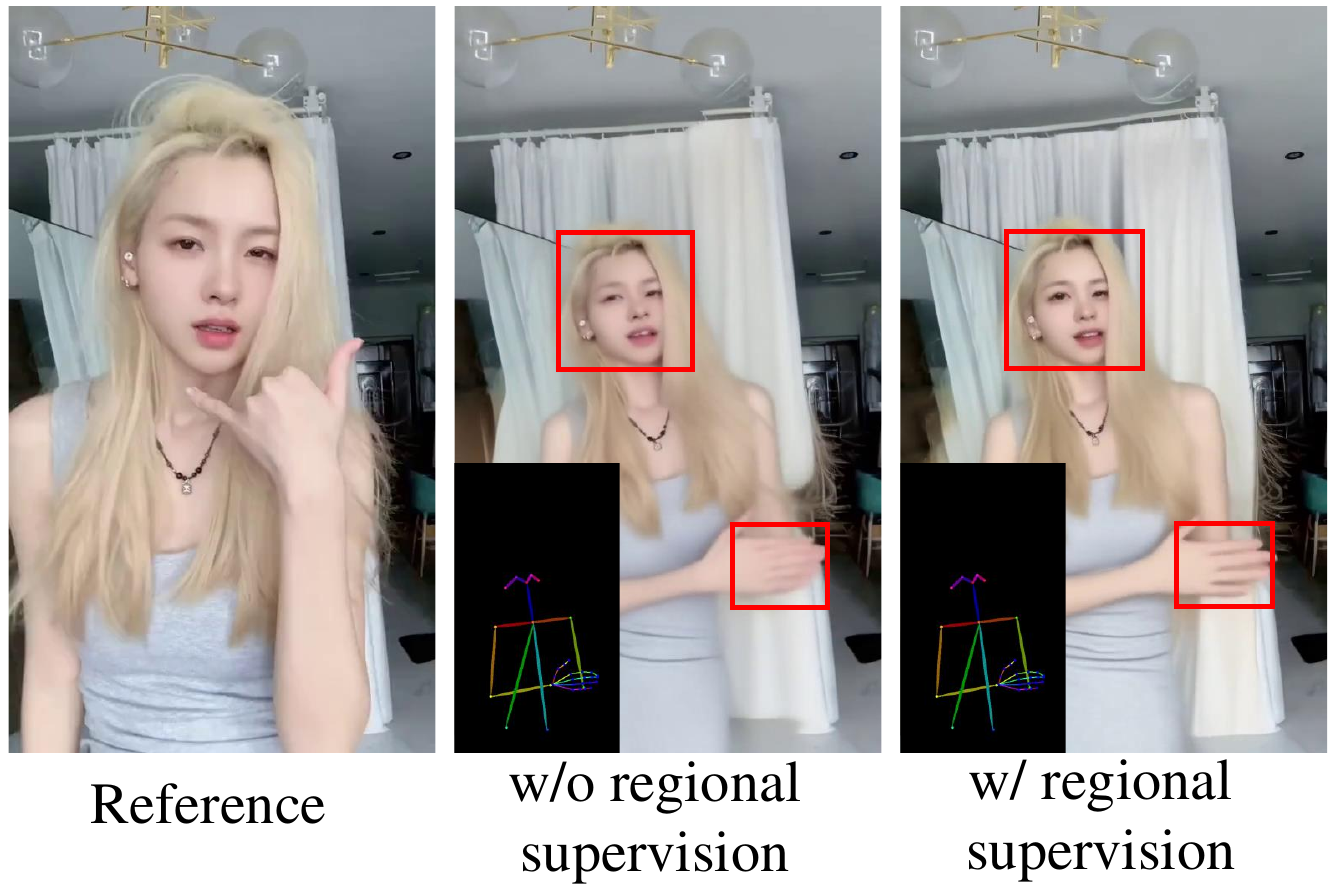}
 \caption{Effects of regional supervision.}
 \label{fig:ab:region}
\end{subfigure}
\hfill
\begin{subfigure}[b]{0.48\linewidth}
 \centering
 \includegraphics[width=\textwidth]{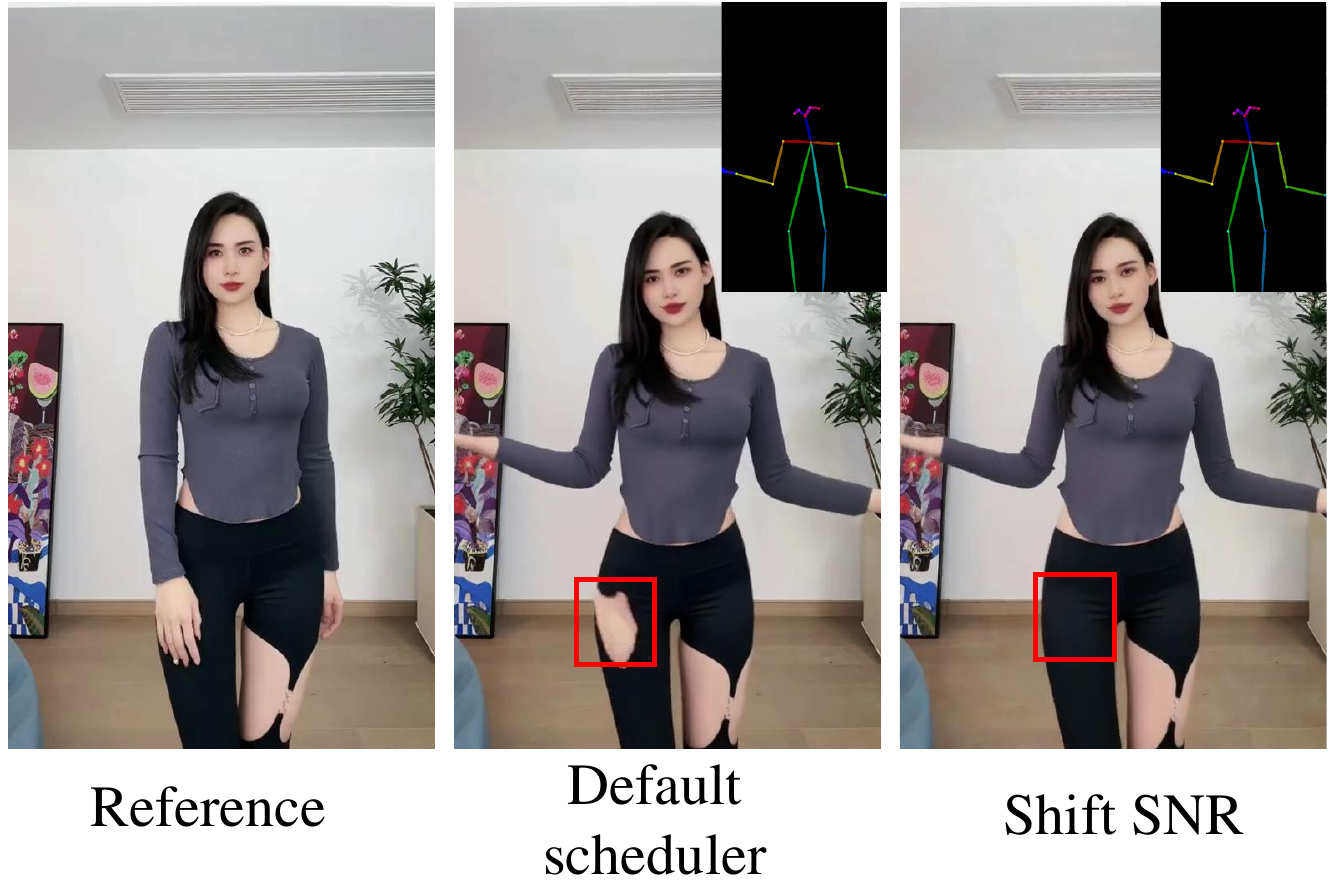}
 \caption{Effects of shift SNR.}
 \label{fig:ab:joint}
\end{subfigure}
\caption{Visualization of ablation studies, with errors highlighted in red boxes. Each frame includes an overlay of the target pose in the bottom left or top right corner for reference.}
\label{fig:ab}
\end{figure}

\begin{figure}[!h]
\centering
\includegraphics[width=\textwidth]{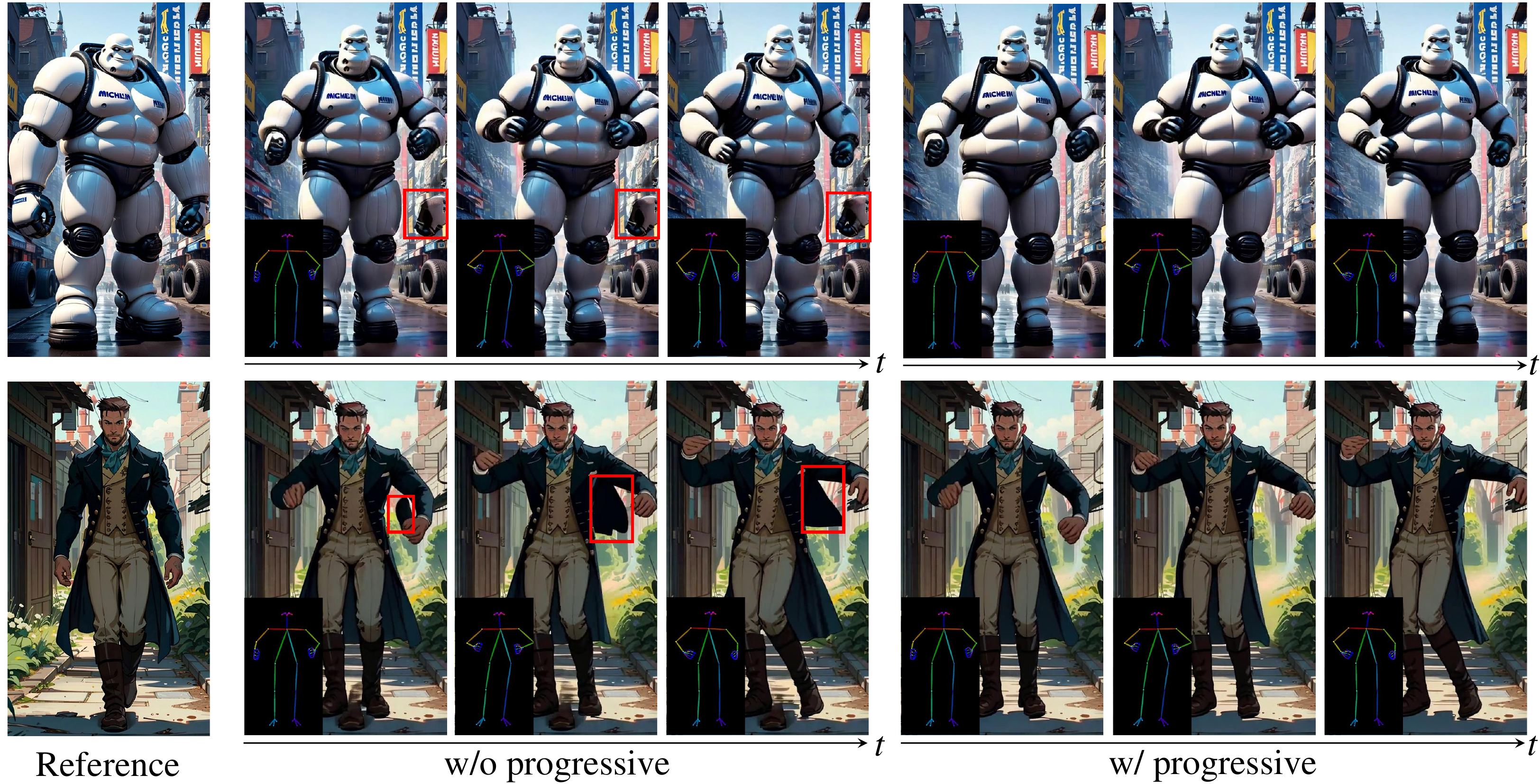}
\caption{
Effects of progressive training. Without progressive training, our model fails to transfer the reference image into the target pose accurately, resulting in artifacts in the background, as highlighted in the red boxes.
}
\label{fig:ab:progressive}
\end{figure}

To verify the design choices in our method, we conduct ablation studies on the HumanDance dataset. The quantitative results for training and inference techniques are shown in Table~\ref{tab:ablation_train} and Table~\ref{tab:ablation_infer}, respectively. Additionally, we present the qualitative results for the regional supervision stage, training strategy, and progressive training.

\textbf{Regional supervision.} 
To evaluate the importance of regional supervision, we remove this stage during training. The results without the regional supervision stage are presented in row 1 of Table~\ref{tab:ablation_train}, showing that most metrics improve when regional supervision is included. We present the qualitative comparison results in Figure~\ref{fig:ab:region}. Human faces are better preserved, and hands are clearer after incorporating the regional supervision stage.

We also validate the choice of fine-tuning only the appearance encoder in the regional supervision stage rather than fine-tuning all spatial modules (\ie, UNet, ControlNet, and appearance encoder). The results of fine-tuning all spatial modules in the regional supervision stage are shown in row 2 of Table~\ref{tab:ablation_train}. All metrics degrade when we fine-tune all spatial modules, performing even worse than when this stage is removed.

\textbf{Motion blur condition.} 
We also evaluate the impact of the motion blur condition by removing it and using only the openpose keypoint sequence as our driving signal. The results without the motion blur condition are shown in row 3 of Table~\ref{tab:ablation_train}. It shows that adding the motion blur condition provides benefits across all metrics. 

\textbf{Training strategies.} 
We then verify the effectiveness of our training strategies, such as shifted SNR and progressive training. To evaluate the impact of removing shifted SNR, we use the default noise scheduler instead (results in row 4 of Table~\ref{tab:ablation_train}). Using shifted SNR proves to be more effective than the default noise scheduler when training at high resolutions like 512$\times$896. The qualitative results in Figure~\ref{fig:ab:joint} also support our quantitative observations.
To study progressive training, we train the temporal module at full resolution without the half-resolution training phase. The results are shown in Figure~\ref{fig:ab:progressive}. Our model generates artifacts in the background when removing the progressive training. 

\textbf{Inference techniques.} As introduced in Sec.~\ref{sec:inference}, we apply three inference techniques in our method. Row 1 of Table~\ref{tab:ablation_infer} shows a simplified version of our model without animation-cfg (A-cfg), prompt traveling (PT), or initial reference noise (IRN). In rows 2-4, we show the effects of individually disabling these inference techniques from the full model. We observe that the use of initial reference noise (row 4) yields the most significant quantitative improvement, followed by prompt travel (row 3) and animation-cfg (row 2).
\newpage
\section{Conclusion}
This work introduces \ours{}, a high-quality diffusion-based human image animation framework. Through the integration of regional supervision, \ours{} enhances identity preservation for human faces and improves fidelity in small-scale regions like the face and hands. Additionally, by adopting an explicit motion blur condition, \ours{} accurately models motion blur and synthesizes animation results closer to the ground truth distribution. Leveraging shifted SNR and a progressive training strategy, our model generates high-fidelity animations with improved generalization ability for unseen domain samples. Experimental results demonstrate that \ours{} outperforms state-of-the-art approaches, achieving significant improvements in reconstruction precision and perceptual quality.

\bibliography{Reference}

\newpage
\appendix
\section{Appendix}
\subsection{Project page}
We include a project page in the supplementary material, please uncompress the project\_page.zip and open index.html for visualization of our video results.

\subsection{Details for evaluation metrics}
\label{appendix:eval}
We follow the prior work DisCo and adopt its evaluation codebase\footnote{https://github.com/Wangt-CN/DisCo} to compute evaluation metrics. Notably, this codebase has an issue with the computation of PSNR metrics, and we use the corrected version for PSNR. Unlike DisCo, which resizes image resolution, we maintain the training resolution, i.e., 512$\times$896, for computing L1 error, PSNR, and SSIM. Additionally, for FID, FID-VID, and FVD computation, we pad the video frames to square dimensions.

\subsection{Details for HumanDance dataset}
We collect the HumanDance dataset from online video sources of social media and get 3552 video clips in total. We additionally mix UBC~\cite{dong2019towards} dataset with the online data for training. Each video has a duration of 15$\sim$20s. For evaluation, we reserved 50 videos for testing purposes and utilized the remaining 3802 videos for training.

\subsection{Dataset preprocessing pipeline}
\label{appendix:process}
We follow a specific pipeline to process these datasets, as outlined below:
\begin{enumerate}
    \item \textit{Keypoint Estimation}: A keypoint estimation model named RTMPose~\cite{jiang2023rtmpose} is used to detect full body keypoints. We empirically find that this estimation model is not robust for feet, we therefore utilize DWPose~\cite{yang2023effective} to estimate feet and merge the keypoints with RTMPose.
    \item \textit{Motion Blur Condition}: We first compute hand movement vector based on the keypoints of two consecutive frames. Second, we crop the hand images based on the keypoints and then calculate variance of Laplacian operator to get the sharpness score.
\end{enumerate}
To augment the training data, we flip the images and motion sequences horizontally.

\subsection{Implementation details}
\label{appendix:impl}
We implement our method using PyTorch, and optimized it using the Adam optimizer. We use a batch size of 32 with gradient accumulation of 4 steps for spatial stages and a batch size of 8 for temporal stages. Our model is trained on 8 Nvidia A100 GPUs. Our training process consists of four stages: (1) spatial training for 35000 iterations (70 hours); (2) finetuning appearance encoder with regional supervision for 4000 iterations (8 hours); (3) half-resolution temporal training for 20000 steps (24 hours); (4) full-resolution temporal training for 1000 steps (2 hours).

\subsection{Details for baselines}
We reproduce the training process for MagicAnimate based on the inference codebase\footnote{https://github.com/magic-research/magic-animate} released by the authors. For AnimateAnyone, we employ the codebase and settings released by the third-party developers\footnote{https://github.com/MooreThreads/Moore-AnimateAnyone}. As for Champ, we directly use their official implementation\footnote{https://github.com/fudan-generative-vision/champ} and settings. 

\subsection{Additional ablation study results}
\begin{figure}[t]
\centering
\begin{subfigure}[b]{0.48\linewidth}
 \centering
 \includegraphics[width=\textwidth]{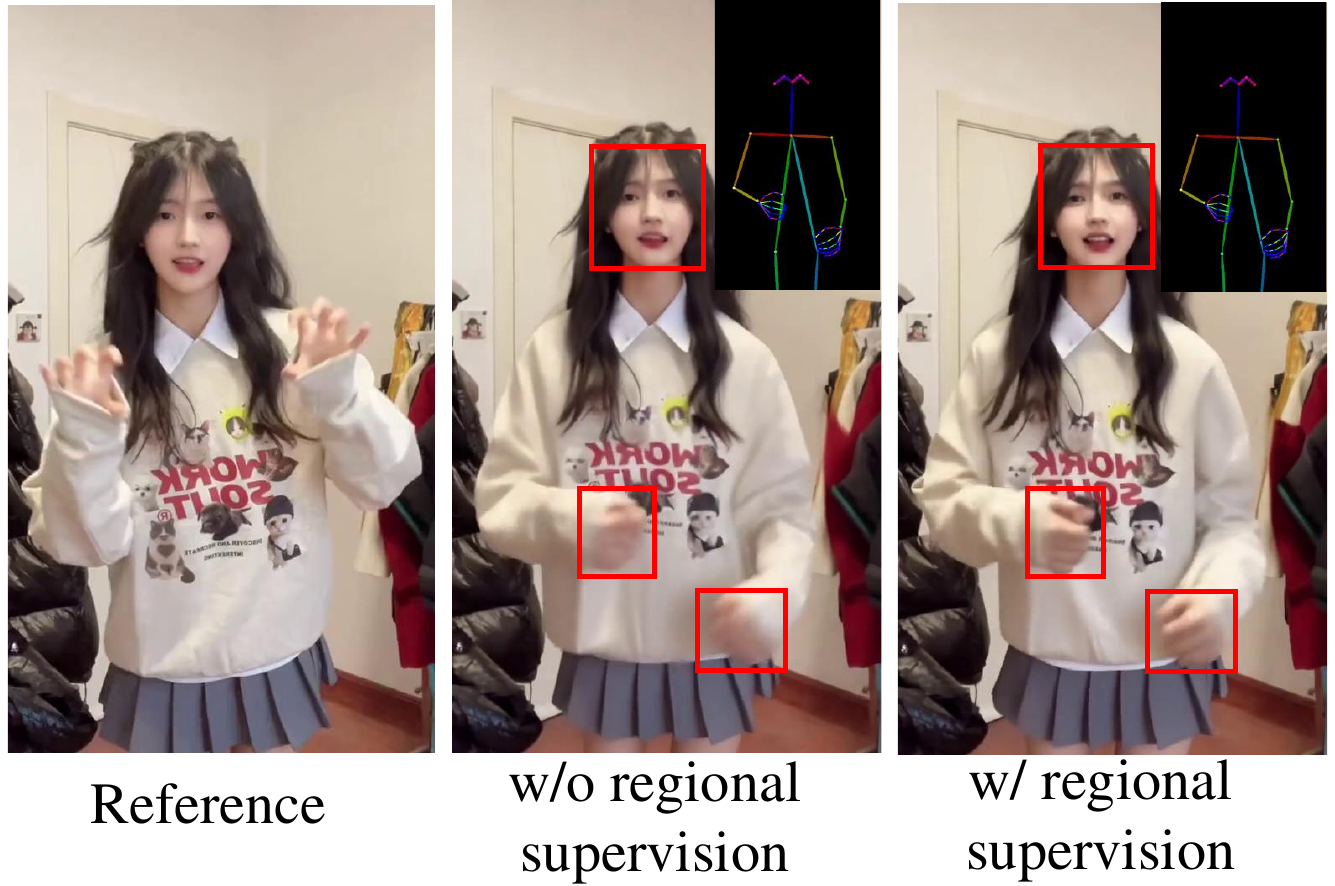}
 \caption{Effects of regional supervision.}
 \label{fig:ab:region_supp}
\end{subfigure}
\hfill
\begin{subfigure}[b]{0.48\linewidth}
 \centering
 \includegraphics[width=\textwidth]{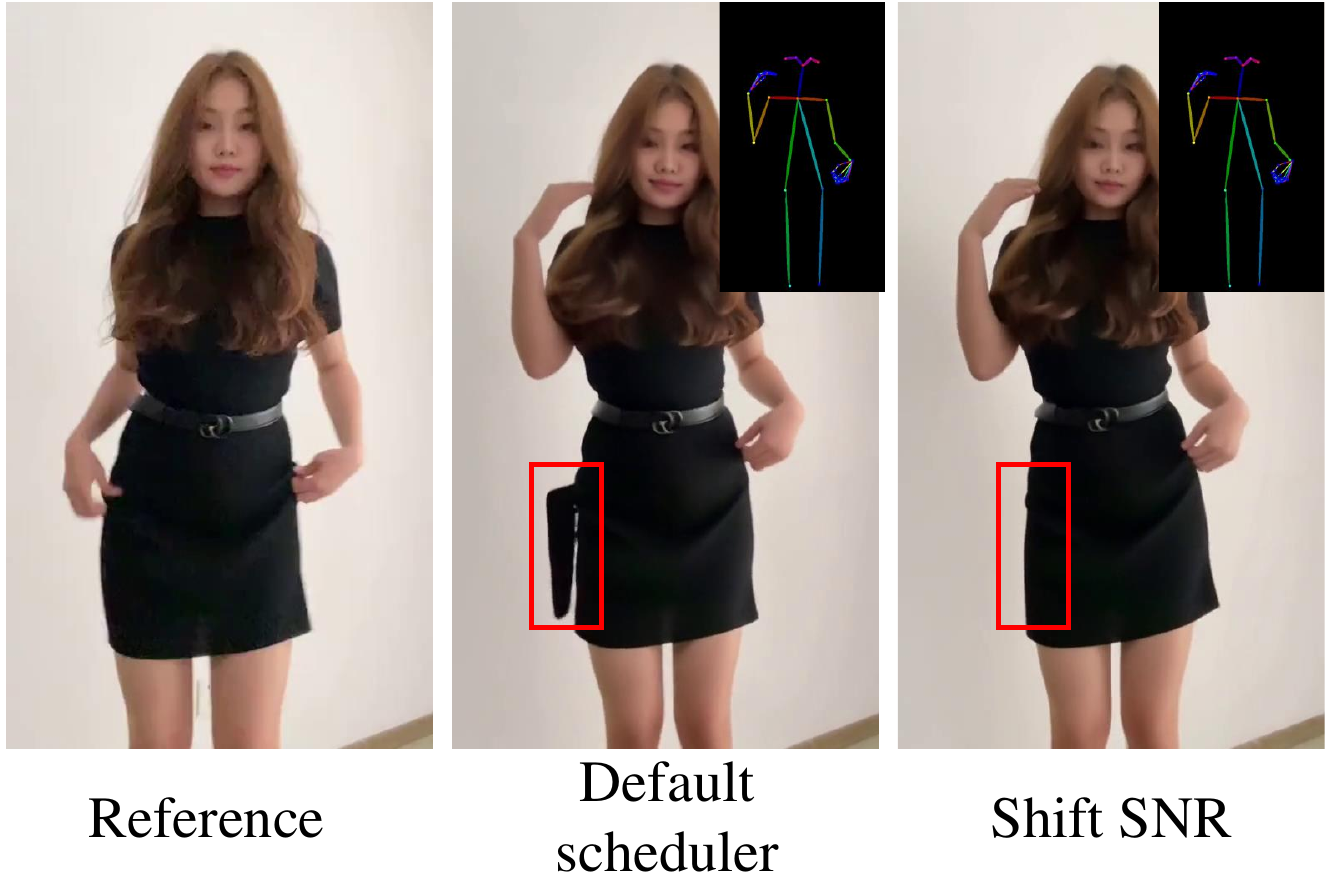}
 \caption{Effects of shift SNR.}
 \label{fig:ab:joint_supp}
\end{subfigure}
\caption{Visualization of ablation studies, with errors highlighted in red boxes. Each frame includes an overlay of the target pose in the top right corner for reference.}
\label{fig:ab_supp}
\end{figure}

\begin{figure}[t]
\centering
\includegraphics[width=\textwidth]{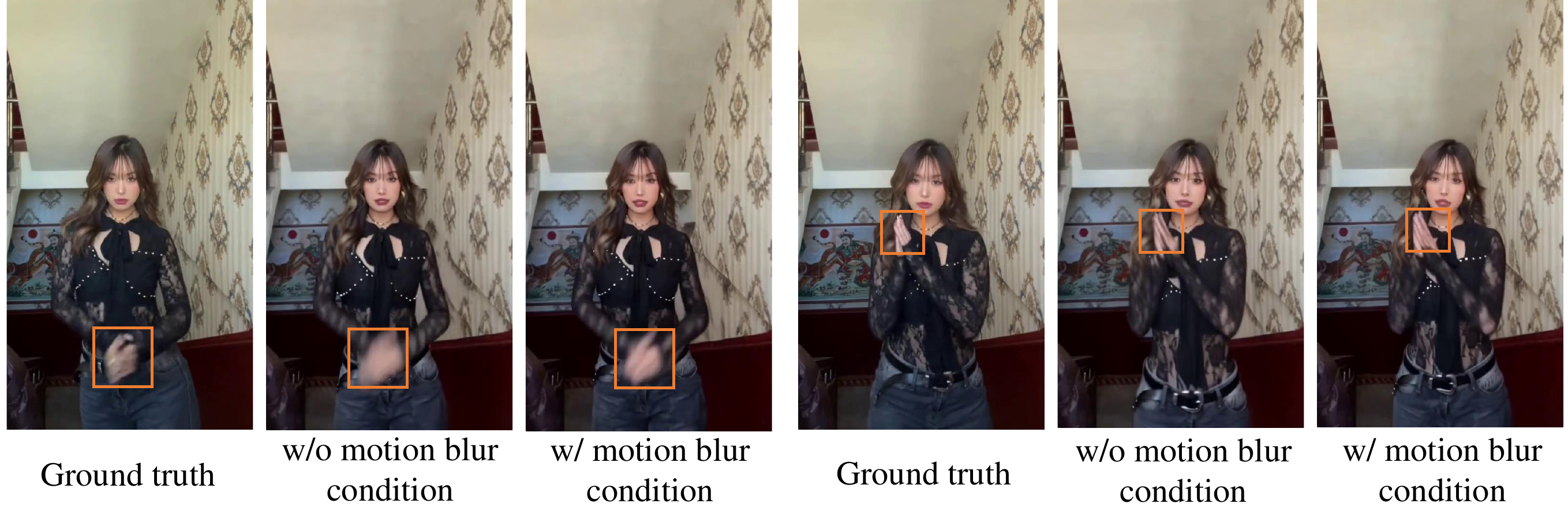}
\caption{
Effects of motion blur condition. Without motion blur condition, our model synthesizes video frames with blurry hands randomly.
}
\label{fig:ab:motion_blur}
\end{figure}

In this section, we extend our ablation studies on regional supervision and shift SNR to include more samples, as shown in Figure~\ref{fig:ab_supp}. We also demonstrate qualitative ablation study results on motion blur condition in Figure~\ref{fig:ab:motion_blur}, where the human hands' clarity is similar to ground truth with the motion blur condition.

\section{Limitations}
\label{limit}
Although \ours{} enables high-quality human image animation, there still exists improvement room in our framework: (1) The accuracy of the control signal estimation method is critical for the precision and robustness of human image animation. Though 2D keypoints are significantly more accurate than other human pose types, like SMPL and DensePose, it is not perfect. We believe a more accurate keypoint estimator would benefit this task. (2) The 2D keypoints cannot convey any 3D prior, which leads to obvious distortion when the motion sequences contain actions like rotation. Incorporating 3D human priors would be helpful to alleviate this issue. (3) Though StableDiffusion contains visual priors for image generation, which could support the inpainting of missing parts in human avatar animation, its capability for hand generation is limited. It is worth exploring a stronger base UNet model for improving hand fidelity.

\section{Broader impact}
\label{appendix:broader_impact}
Our human image animation method could be misused for harmful purposes such as fraud or harassment. These malicious applications may pose a societal threat.

The datasets used to develop our model have unbalanced demographic distributions. Consequently, one must bear this in mind when deploy the model considering the fairness issues.

We implement safeguards and protect our model from misuse by applying license agreements for model download and usage. We believe this rule can add restrictions on the access to our model.

\section{Reproducibility}
\label{appendix:reproduce}
In this supplementary material, we provide comprehensive information to ensure the reproducibility of our work. We introduce the implementation details (Section~\ref{appendix:impl}), dataset pre-processing pipeline (Section~\ref{appendix:process}), and details for evaluation metrics (Section~\ref{appendix:eval}).



\end{document}